# 基于预训练语言模型的 BERT-CNN 多层级专利分类研究*

陆晓蕾[1]　倪斌[2]

1. 厦门大学外文学院　厦门　361000; 2. 深湃信息科技（深圳）有限公司　深圳　518000;)

**摘要**：专利文献的自动分类对于知识产权保护、专利管理和专利信息检索十分重要，构建准确的专利自动分类器可以为专利发明人、专利审查员提供辅助支持。本文以专利文献分类为研究任务，选取国家信息中心公布的全国专利申请信息为实验数据，提出了基于预训练语言模型的 BERT-CNN 多层级专利分类模型。实验结果表明，在该数据集上，BERT-CNN 模型在准确率上达到了 84.3%，大幅度领先于卷积神经网络和循环神经网络等其他深度学习算法。本论文的局限性在于未将模型扩展到更深层级的分类中。通过 BERT 抽取的向量在表达词汇与语义方面具有强大的性能，在专利自动分类领域具备实际应用效果。

**关键词**：专利；文本分类；深度学习；BERT

**分类号**：G254.1

# BERT-CNN: a Hierarchical Patent Classifier Based on a Pre-Trained Language Model

Lu Xiaolei[1]　　Ni Bin[2]

1. School of Foreign Languages and Cultures, Xiamen University, Xiamen 361000; 2. DeepAI Tech Co., Ltd, Shenzhen 518000;

**Abstract**：The automatic classification is a process of automatically assigning text documents to predefined categories. An accurate automatic patent classifier is crucial to patent inventors and patent examiners in terms of intellectual property protection, patent management, and patent information retrieval. We present BERT-CNN, a hierarchical patent classifier based on pre-trained language model by training the national patent application documents collected from the State Information Center, China. The experimental results show that BERT-CNN achieves 84.3% accuracy, which is far better than the two compared baseline methods, Convolutional Neural Networks and Recurrent Neural Networks. We didn't apply our model to the third and fourth hierarchical level of the International Patent Classification — "subclass" and "group".The visualization of the Attention Mechanism shows that BERT-CNN obtains new state-of-the-art results in representing vocabularies and semantics. This article demonstrates the practicality and effectiveness of BERT-CNN in the field of automatic patent classification.

**Keywords**：　Patent; Text Classification; Deep Learning; BERT

# 1　引言

近年来，工业界和学术界产生了大量专利申请。据世界知识产权保护组织（WIPO）的指导

性报告显示：2017 年，全球所有专利局共受理的专利申请超过三百万个，并且这项数据每年还会以更快的速度增长。出于检索和管理需要，专利文献需按照专业技术领域进行分类与赋码。1971 年《斯特拉斯堡协定》提出的国际专利分类法（IPC）是国际上通用的专利文献分类法，几乎覆盖了所有科技领域，超过 90 个国家使用该标准。该分类标准提供了一种由独立于语言的符号构成的树形体系，按所属技术领域对发明专利和实用新型专利进行分类。IPC 分类包含"部（section）-类（class）-亚类（subclass）-组（group）"四个层级，其中"组"级共含有 74503 种类别（WIPO，2019）。过去，传统的人工分类受限于专利员的自身知识与经验，难以做到快速和精确。随着专利申请数量的剧增，IPC 分类法的多层级、多标签等特点给专利审查人员带来了巨大的工作量。同时，快速的全球化竞争以及完善的知识产权保护体系对文献分类技术提出了更高的要求。随着深度学习和文本分类的快速发展，越来越多的研究开始聚焦于专利文献的自动分类。目前，面向自然语言处理的专利分类研究多集中于英文专利申请，然而，中国的专利申请数量占全球比重达 43.6%[1]，构建一个相对精确的中文专利分类器的需求也日益凸显。本文试图利用基于卷积神经网络、循环神经网络和 BERT 预训练模型等深度学习手段建立中文专利文献分类器，提出了 BERT-CNN 模型，并探索多层级分类方法，为快速准确的专利文本分类提供参考。

## 2 相关研究

过去，大量的专利文献自动分类研究主要集中于朴素贝叶斯[2]、支持向量机[3]和 K-近邻算法[4]等统计方法。Fall 等[5]对这些算法进行了比较评估，阶段性地总结了这三种分类器的效果，提出了支持向量机是当时最好的分类器。D'hondt 等[6]从多元短语组等统计语言学的角度对专利文档进行了分析，在一定程度上提高了专利分类的准确性。姜春涛等[7]建立基于图结构的文本分类器对四个不同技术领域的专利进行分类。然而，以上方法大都需要人工特征工程，建模过程费时费力，准确性也存在瓶颈。随着神经网络模型在自然语言处理上的广泛应用，尤其是以词向量为代表的文本表征技术的进步[8]，文本分类的效果得到了显著提升。Grawe 等[9]利用专利文献训练 Word2Vec 词向量模型，在此基础上仅叠加一层 LSTM 作为分类器，获得了较好的分类效果。类似的，Shalaby 等[10]基于 LSTM 训练固定层次结构向量（Fixed Hierarchy Vectors）对文档表示进行了增强学习，极大提升了分类的准确性。可见，在专利文档分类的研究中，词或文档表示的效果好坏是构建一个快速并且稳定的分类器的关键。

传统的词向量，如 Word2Vec[8]和 Glove[11]，主要通过建立语言模型[12]获得，在表达多义词方面存在较大缺陷。为此，Peters 等[13]提出的 Elmo 模型利用双向长短期记忆网络（Bi-LSTM）生成词的上下文表示，在获得预先训练的词向量之后根据实际数据的上下文进行向量表达的动态调整。Alec 等[14]提出基于 Transformer 的生成式预训练词向量模型（Generative Pre-Training，GPT），在多项 NLP 任务中获得了当时的最高分数。Devlin 等[15]吸收了 Elmo 和 GPT 模型的优势，提出了基于 Transformer 的双向编码器模型（BERT），凭借其优秀的表达词句能力，在 GLUE 排行榜上，刷新了多达 11 项 NLP 任务的记录。BERT 预训练模型可以根据具体下游任务进行参数微调。在此基础上，通过简单的接口处理其它自然语言处理任务，如文本分类。目前，BERT 已经广泛应用于如命名实体识别[16]、阅读理解问答[17]等各项自然语言处理任务中，自然语言处理自此进入了预训练模型的大规模应用新阶段。虽然 BERT 提供了下游任务的简单接口，可以直接进行文本分类。然而，BERT 作为预训练语言模型，关于其作为文档向量的研究和应用尚不多见。过去，fastText[18][19]通过将整篇文档的词及 n-gram 向量叠加平均得到文档向量，然后使用文档向量直接做 softmax 多分类，在处理简单文本分类中获得了较好的效果。但是这种利用词向量（word embedding）表达文本的方式无法区分多义词。而 BERT 在预训练的基础上，根据具体语句进行参数微调，在某种程度上缓解了这一问题。因此，本文提出了 BERT-CNN 模型，尝试将 BERT 作为编码器，将专利摘要映射成文档向量，在其上叠加更加复杂的卷积神经网络。

# 3 研究方法

## 3.1 BERT-CNN 模型结构

本文基于 BERT 预训练语言模型,提取其顶部四层 Transformer 的输出作为专利摘要的文档向量,与卷积神经网络进行联合训练(图 2)。

### 3.1.1 BERT 层

BERT 的英文全称是 Bidirectional Encoder Representations from Transformers,采用双向 Transformer 编码器[20],利用多头注意力机制融合了字左右的上下文信息。同时,与早期通过训练语言模型的目标任务——"预测下一个词"[12]不同的是,BERT 设置了两种目标任务分别获取单词与句子级别的表义方式:1)遮盖语言模型;2)上下句关系预测。其中,遮盖语言模型类似"完形填空",即随机遮盖 15%的句子,让编码器预测这些词;上下句关系预测通过预测两个随机句子能否组成上下句来学习句子间的关系。通过这样训练出来的模型,BERT 具有很强的句词表达能力,无论是在字词级别的 NLP 任务,如命名实体识别,还是在如问答类的句子级别的 NLP 任务中,都具有卓越的表现。在 BERT 的顶层,可以直接叠加简单的线性模型,结合具体的任务(specific-task)做参数微调(fine-tune)来完成其他 NLP 任务,如文本分类。

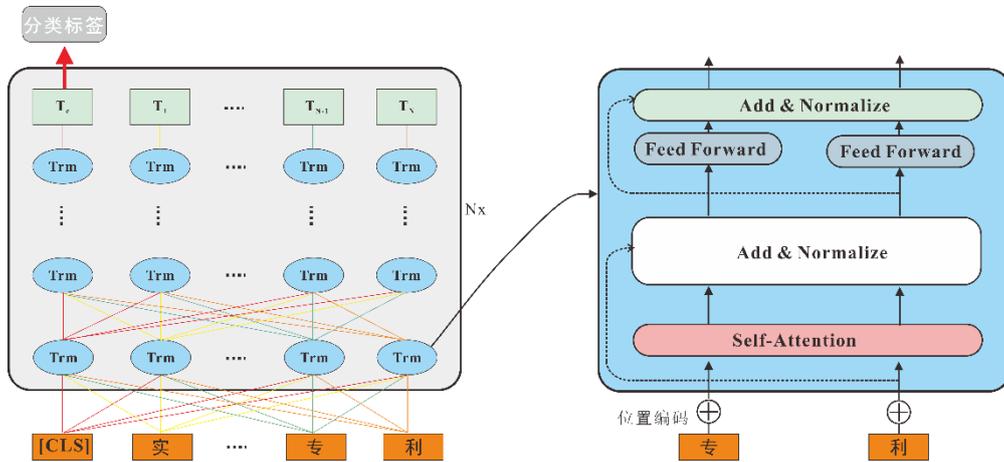

图 1 BERT 网络架构(Nx 为 Transformer 的层数)

BERT 模型的详细结构见图 1,由多层双向 Transformer 构成,每个 Transformer 利用多头自注意力机制(Multi-Head Attention)建立词与词之间的联系强弱(权重)。谷歌开源了两种不同规模的 BERT 模型,分别为 BERT-Base 和 BERT-Large[15]。其中,BERT-Base 拥有 12 个 Transformer 层,768 个隐含单元,12 个自注意力层,总共含有 1.1 亿个参数;BERT-Large 拥有 24 个 Transformer 层,1024 个隐含单元,16 个自注意力层,总共含有 3.4 亿个参数。由于计算资源限制,一般研究者多采用前者,本文亦选取 BERT-Base 作为预训练模型。BERT-CNN 采用 BERT 中的后四层 Transformer 层的输出作为下游 CNN 模型的输入。

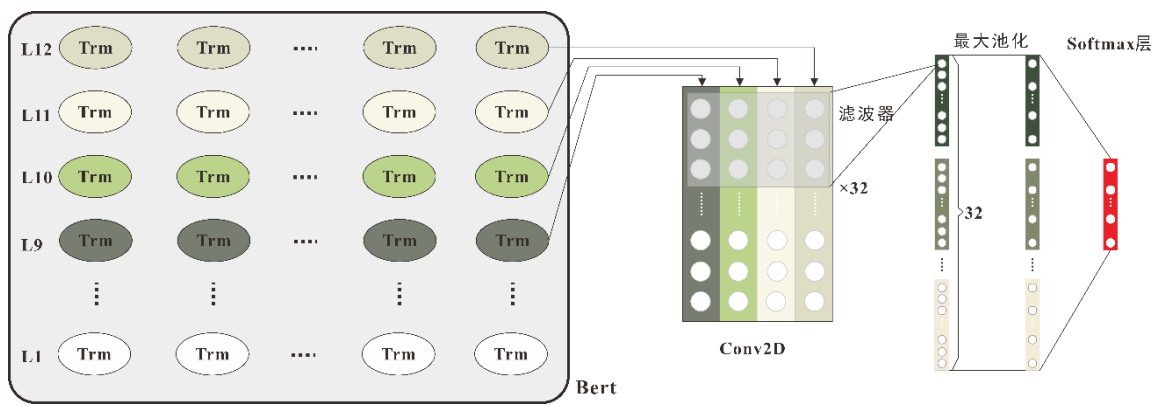

图 2 BERT-CNN 模型结构图

### 3.1.2 Conv2D 层

卷积神经网络（CNN）在大部分情况下用于图像识别，但是在自然语言处理中也获得了很好的效果[21]。BERT 中每一层 transformer 的输出都可以作为句向量（维度为 768 的向量），本文取其最后四层 $L_{12}$、$L_{11}$、$L_{10}$、$L_9$ 作为 CNN 的输入矩阵 I（768×4）。然后用 32 个滤波器 F（3×4），步长为 1，扫描输入矩阵 I，目的是提取文本 3-Gram 特征，通过 I⊗F 内积获得 32 个特征向量。为了降低计算的复杂度，CNN 通常使用池化计算降低矩阵的维度。池化计算有最大池化（Max-Pooling）与平均池化（Average-Pooling）两种方式，一般采用最大池化，即在池化窗口中选取最大的元素。由于专利文本分类对于局部的某些关键词或术语比较敏感，本文选取最大池化方式。经过最大池化层后拼接并通过 Softmax 层获得专利分类的概率分布。在网络训练过程中，滤波器的参数是不共享的，并且与网络间的连接参数同时更新。

## 3.2 多层次分类架构

传统的分类方法在处理多层级任务时没有区分层级，将其视为普通的多分类（multi-class）任务，这在处理类别有限的任务时简单有效。但是，随着层级和类别的增加，其效果表现越来越差。尤其是在图书分类或者专利分类等多层级任务中，类别数量庞大（如 IPC 中"组"的分类数高达 74503 种），算法的准确率随着类别的增加急剧下降。另外，传统分类方法将所有类别独立看待，在某种程度上忽视了类别之间的联系。例如，在 IPC 分类中，"A21C3/00"与"A21C7/00"在专业领域上很接近，而与"D21C7/00"相去甚远。然而，这种方法将三者看成独立的类别，没有任何联系，这显然是不恰当的。

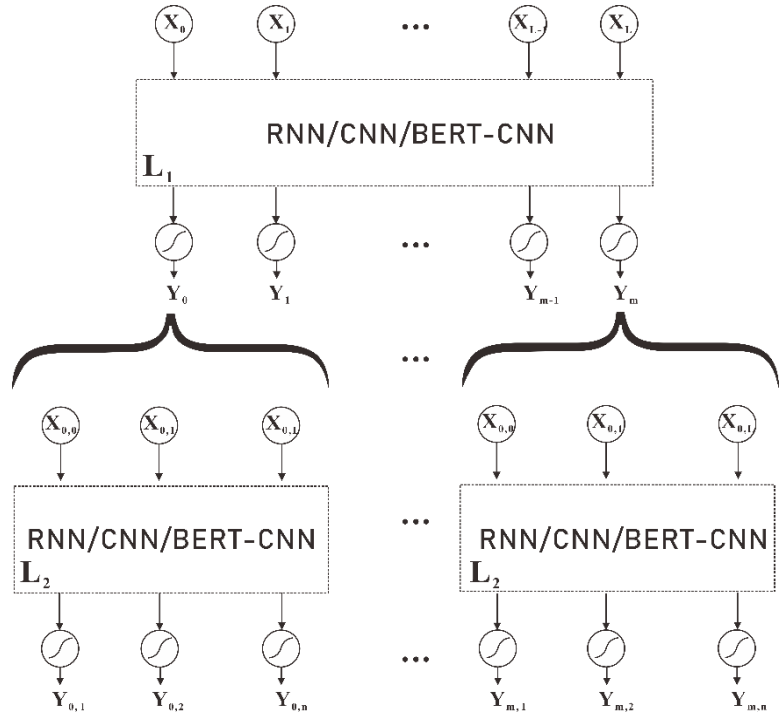

图 3 本文的多层级分类网络架构

针对专利分类的多层级问题，本文提出了一种堆叠式的神经网络组合训练模型（图 3）。模型分为两层 $L_1$ 和 $L_2$，其中 $L_1$ 为一级分类层，$L_2$ 为二级分类层。模型首先获取所有训练样本输入，经过不同的分类器（本文中采用 RNN/CNN/BERT-CNN）后训练一级分类 $Y_{L1}$。随后，在中间设置数据转换层。该层的作用是根据 L1 层的标签筛选 L2 层网络。具体做法是根据 $L_1$ 层的标签值 $Y_m$ 获取所有一级分类为 $Y_m$ 的样本作为 $L_2$ 层的训练样本。例如，若 $L_1$ 层输出为'$Y_2$'，则 $L_2$ 层将只获取所有一级分类为'$Y_2$'的数据进行训练。

## 4 试验与结果

### 4.1 数据集

本文采用国家信息中心提供的全国专利申请数据[1]。数据总量达到 277 万条。时间跨度为 2017 年全年（按照专利申请时间统计），地域覆盖全国，数据主要包含共 16 个核心字段：申请号、申请日、公开号、公开日、专利名称、专利分类、摘要、申请（专利权）人、发明人、申请人地址、申请人邮编、所在省代码、所在市代码、所在县代码、法律状态、有效性。数据格式为 CSV，编码格式为 GBK。本实验提取"摘要"和"专利分类"两个字段进行分类模型训练。专列分类采用 IPC 分类法，该分类法将专利分为"部（section）-类（class）-亚类（subclass）-组（group）"四个层级，如"A01D42/04"，"A"是部，"01"是类，"D"是亚类，"42/04"是组。本文为简单起见，仅选取前两层进行分类，即部和类。

---

[1] 国家信息中心, 2017, "发明专利数据", http://dx.doi.org/10.18170/DVN/ASRTHL, 北京大学开放研究数据平台, V2

图 4 预处理后的专利数据集

经过剔除无效数据（分类号为非严格的 IPC 分类）等预处理操作后，原始数据剩余约 231 万条（数据结构见图 4）。该数据集包含了 IPC 的所有部（字母 A 至 H）和类（130 个）。其中，平均每个类含有 17828 条文本。详细的文本与分类统计见表 1。

表 1 专利数据集文本与类别相关统计

| 部名 | 含义（参考国家知识产权局[2]） | 类数 | 文本数 | 每类的平均文本数 |
| --- | --- | --- | --- | --- |
| A | 人类生活必需 | 16 | 327537 | 20471 |
| B | 作业；运输 | 38 | 664602 | 17490 |
| C | 化学；冶金 | 21 | 201984 | 9618 |
| D | 纺织；造纸 | 9 | 37673 | 4186 |
| E | 固定建筑物 | 8 | 155250 | 19406 |
| F | 机械工程；照明；加热；武器；爆破 | 18 | 252325 | 14018 |
| G | 物理 | 14 | 378874 | 27062 |
| H | 电学 | 6 | 299456 | 49909 |
| 总计 | | 130 | 2317701 | 17828 |

通过分析发现该数据集存在如下问题：①各部中含有类的数目和文本数不均。其中，B 部的类别最多，达到 38 个类；H 部类别最少，只有 6 个类。其次，各部和各类文本数量分布不均衡。②部间存在相似性问题。例如"F：照明"、"G：物理"和"H：电学"三者存在较强的相似性，人工也难以准确区分。数据的稀疏性与类别的相似性都会对模型的准确性造成一定影响。面对这种情况，工业界通常会对数据进行增强。本文的目的是比较 BERT-CNN 和其他基线模型的相对性能，这两大问题不对结果产生影响。

### 4.2 实验配置

本文实验代码基于 python 3.7.3，深度学习框架主要利用 TensorFlow r1.13，计算机处理器为 Intel i7 的 12 核处理器，GPU 显卡为 Nvidia 2080Ti，运行内存为 32G。

本实验选取卷积神经网络和循环神经网络作为对照基线模型。在基线系统中，我们使用 300 维 word2vec 中文词向量[22]。文本最大长度 L 设为 200 个词，网络训练的批处理（batch size）大小为 20，Epoch 为 20，学习率 $\eta$ 为 2e-5。BERT-CNN 模型使用中文 BERT_Basic 模型(Chinese_L-12_H-768_A-12)，学习率为 2e-5，batch_size 为 24，最大文本长度 L 为 200，Epoch 为 20，优化器为 Adam。BERT-CNN 中的 CNN 滤波器个数为 32，激活函数为 ReLU。

本文按 9:1 将数据集划分为训练集与测试集。基线方法的数据经过了分词预处理后利用第三

---
[2] http://epub.sipo.gov.cn/ipc.jsp

方训练的词向量。BERT-CNN 模型不需要分词。

### 4.3 结果分析与讨论

#### 4.3.1 评估指标

本文模型使用正确率作为评估指标，最终联合模型的正确率Acc(X)通过以下公式计算：

$$Acc(X) = Acc(L_1) \sum_m \frac{N_j}{M} Acc(L_{2m}) \qquad (1)$$

其中，$Acc(L_1)$代表第一层$L_1$模型的预测正确率，$m$代表第一层分类（部）的数量，$M$代表所有样本的数量，$N_j$表示部$j$含有的二级分类（类）的样本数。$Acc(L_{2m})$表示第二层（类）模型的正确率。

#### 4.3.2 优化器

我们使用 Adam 作为深度学习网络的梯度下降优化器[23]。Adam 结合了 AdaGrad 和 RMSProp 两种优化器的优势，综合考虑了一阶距估计（First Moment Estimation，即梯度的均值）$m_t$和二阶矩估计（Second Moment Estimation，即梯度的未中心化的方差）$v_t$的计算结果得出更新步长。

$$\theta_t \leftarrow \theta_{t-1} - \frac{\alpha}{\sqrt{\hat{v}}+\epsilon}\hat{m} \qquad (2)$$

$$g_{i,t} = \nabla_\theta J(\theta_i, x_i, y_i) \qquad (3)$$

$$m_t = \beta_1 m_{t-1} + (1-\beta_1)g_{i,t} \qquad (4)$$

$$v_t = \beta_2 m_{t-1} + (1-\beta_2)g_{i,t}^2 \qquad (5)$$

其中，$\hat{m} = \frac{m_t}{1-\beta_1^t}$，$\hat{v} = \frac{v_t}{1-\beta_2^t}$。Adam 适用于不稳定目标函数，在大规模的数据及参数场景中表现出众。由于 Adam 克服了梯度稀疏的问题，实现简单，计算高效，是深度学习中默认较为优秀的优化器。

#### 4.3.3 实验结果

表 2 基于预训练模型的 BERT-CNN 与基线方法的实验结果（准确率%）

| 模型 | Acc($L_1$) | Acc($L_2$) | | | | | Acc($L_{2总}$) |
|---|---|---|---|---|---|---|---|
| RNN | 80.9 | A | 91.8 | B | 79.5 | 86.3 | 69.8 |
| | | C | 89.2 | D | 84.7 | | |
| | | E | 88.1 | F | 88.6 | | |
| | | G | 90.2 | H | 92.4 | | |
| CNN | 81.7 | A | 92.4 | B | 78.0 | 85.8 | 70.1 |
| | | C | 87.5 | D | 85.3 | | |
| | | E | 87.3 | F | 88.7 | | |
| | | G | 85.9 | H | 91.3 | | |
| 本文模型 BERT-CNN | **90.5** | A | 96.5 | B | 89.6 | **93.1** | **84.3** |
| | | C | 94.1 | D | 92.7 | | |
| | | E | 95.2 | F | 93.8 | | |
| | | G | 92.3 | H | 95.7 | | |

表 2 是我们的实验结果，$L_1$ 表示模型预测"部"的准确性，$L_2$ 表示预测单独各"类"的准确性，$L_{2总}$ 表示预测第二层"类"的准确性，是通过计算 $L_2$ 的加权和得出的。实验基线系统选取常用的卷积神经网络与循环神经网络分类器。在该数据集上，BERT-CNN 模型的分类效果最好，在单独层的分类中的准确率基本都超过了 90%，总体分类准确率达到了 84.3%，与 CNN、RNN 相比提升了 14%左右。这种大幅度的提升进一步证实了基于预训练模型的 BERT-CNN 在文本分

类方面的强大性能。另外，可以看出 CNN 和 RNN 在该数据集上分类性能相仿，在面对大数据量和类别较多的情况下还会存在准确率下降的问题。例如，基线方法在处理部级分类时比类级分类的准确率低 4%，显示出在大数据量上的性能下降。同时两种基线方法在类别较多的 B 类上准确率衰减比 BERT-CNN 较为严重。实验结果进一步表明本文模型在处理大数据量和多类别上的鲁棒性。

#### 4.3.4 讨论

上述的结果已经表明，使用基于预训练模型的 BERT-CNN 在提高专利分类准确性方面具有较好的效果。BERT 所提供的文档向量具有良好的字词与语义表征能力，其基于预训练的微调可以有效解决传统词向量一词多义的问题，这是模型获取高准确率的关键。

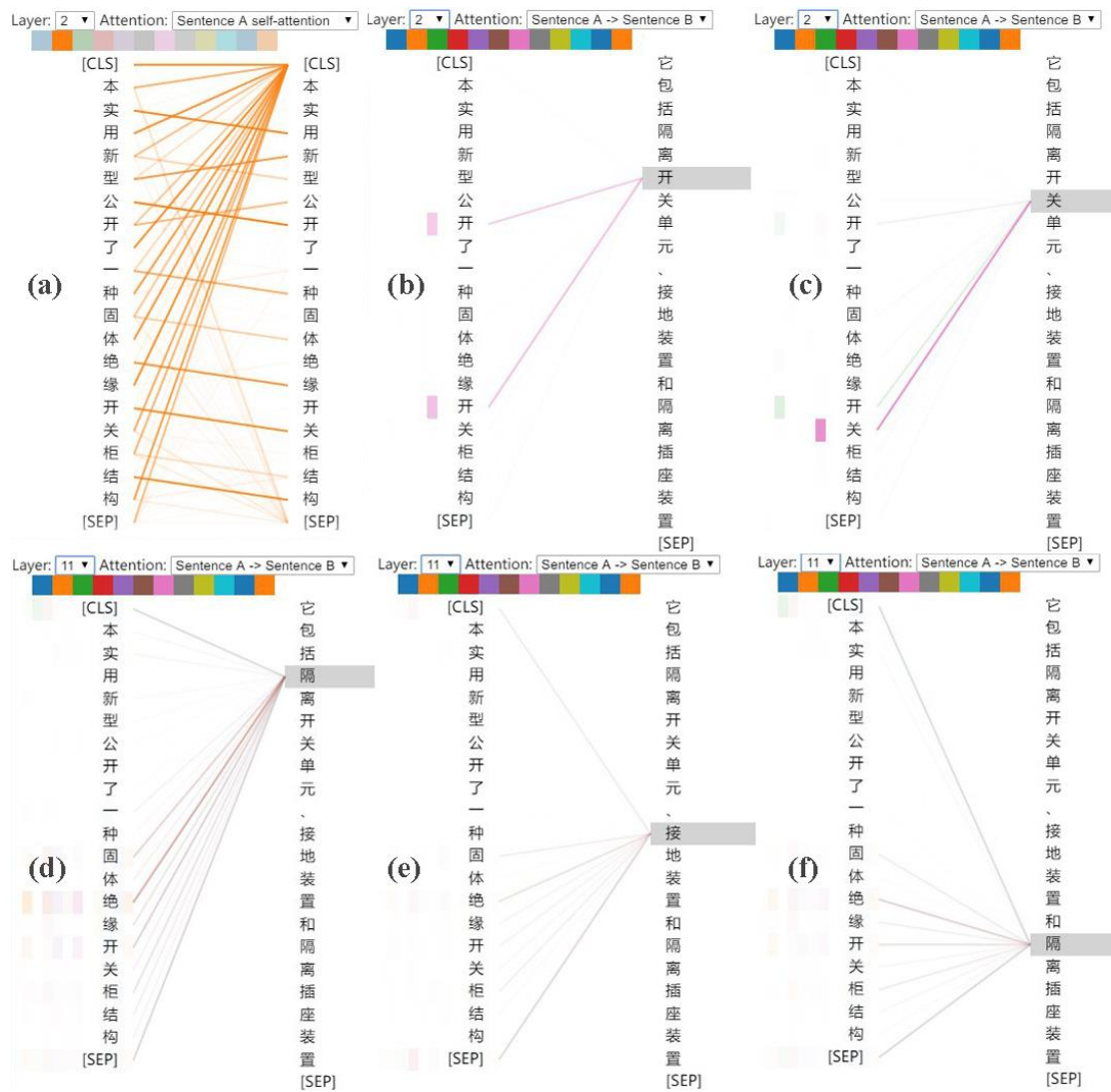

图 5 BERT 模型的注意力机制可视化

BERT 在底层使用双向 Transformer，通过多头注意力机制可以获取字词之间的联系。因此，对注意力的可视化研究可以直观地了解 BERT 模型的注意力机制，帮助解释模型。图 5 利用 Vig 等[24]开源的注意力可视化工具 bertviz 进行模型可视化分析。该工具能用于探索预训练的 BERT 模型各个层以及头部的注意力模式。

以本文的专利分类为例，输入以下两个句子：

句子 A：本实用新型公开了一种固体绝缘开关柜结构。

句子 B：它包括隔离开关单元、接地装置和隔离插座装置。

图 5(a、b、c)和图 5(d、e、f)分别是 BERT 第二层与第十一层的注意力机制的可视化，将注意力可视化为被更新位置（左）和被关注位置（右）之间的连线。线段的颜色代表不同的"注意力头"，线段宽度反映注意力值的大小。BERT 使用[CLS]进行分类任务的第一个输入，[SEP]是句子的分隔。图 5a 显示，在第二层中，注意力主要体现词汇关系：句中的每个字的注意力主要集中在其下一个字，如"实"的注意力集中在"用"。图 5b、c 显示，句间的注意力主要集中在相似的词，如"开"、"关"之间的强烈注意力联系。图 5(d、e、f)显示，在第十一层中，注意力主要体现在语义关系上，如句子 A 中的"固体绝缘开关柜结构"与句子 B 中的"隔离开关单元"、"接地装置"和"隔离插座装置"具有较强的注意力强度。

基于以上发现，我们认为 BERT 中的每一层 Transformer 的输出都可以作为句子或文档向量为其他模型提供输入。具体选取 Transformer 的哪些层作为文档向量在某种程度上也会影响模型的准确率。本文针对该问题，对数据集中的 D 部进行了平行对照实验。从图 6 可以看出，模型的准确率在 Transformer 层数 N=4 时达到最大。当 N<4 时，Transformer 的输出作为文档向量的代表性还不太强，准确率略有下降；当 N>4 时，文档向量中表征词汇语法关系的成分增大，对分类结果意义不大，反而造成干扰和准确率的下降。因此，本文认为利用 BERT 预训练提供的向量与其它模型融合时，应该根据具体的任务特点进行合理选择。

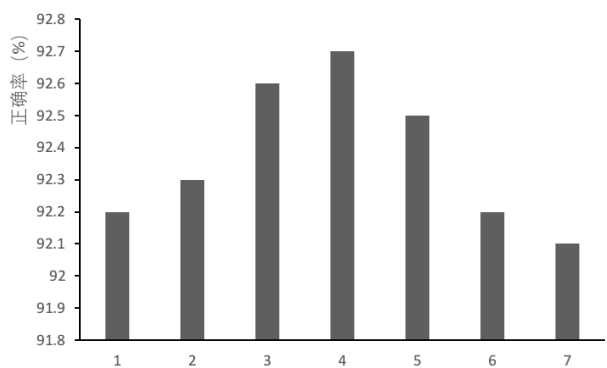

图 6 不同 Transformer 层数作为输入的分类准确率直方图

## 5 结语

文档分类是文献管理与信息处理中的重点和难点。尤其是在专利文献这种多层级的分类领域中，传统的统计方法准确率较低，且建模过程费时费力。本文选取国家信息中心公布的 2017 年全国专利申请数据，以专利的自动分类作为研究目标，在预训练模型 BERT 的基础上抽取文档向量，提出了 BERT-CNN 分层级深度网络分类模型。本文提出的方法在该专利数据集上获得了 84.3%的准确率，大幅度领先过去的卷积神经网络与循环神经网络分类法。实验结果表明，基于预训练语言模型的 BERT-CNN 提取的向量表示，在应用上效果优于传统的词向量。同时，在处理数据量和类别较多的任务上，该模型具备稳定性强的特点。进一步实验表明，Transformer 层数的选取对模型的准确率有一定的影响，应该根据具体的任务特点选择合适的输出层。

本论文的局限性在于未将模型扩展到更深层级的分类中。在后续研究中，将扩展模型层级到"组"，以应用于实际的专利自动分类中；同时，探索更多的预训练模型（例如 GPT-2）来进一步丰富专利文献分类领域的研究。